\documentclass{article}

\PassOptionsToPackage{numbers, compress}{natbib}
\usepackage{amsmath} 
\usepackage{booktabs}
\usepackage{makecell}
\usepackage[preprint]{neurips_2026}
\usepackage{multirow}
\usepackage{graphicx}
\usepackage{xspace}
\usepackage[utf8]{inputenc} 
\usepackage[T1]{fontenc}    
\usepackage{hyperref}       
\usepackage{url}            
\usepackage{booktabs}       
\usepackage{amsfonts}       
\usepackage{nicefrac}       
\usepackage{microtype}      
\usepackage{xcolor}         
\usepackage{algorithm}
\usepackage{algpseudocode}
\usepackage{amsmath}
\usepackage{amssymb}
\usepackage{caption}
\usepackage{float}

\newcommand{\methodname}{WorldDirector}
\newcommand{\method}{\texttt{\methodname}\xspace}

\newcommand{\tocite}[1]{{\color{red} [TO CITE]}}

\algrenewcommand\algorithmiccomment[1]{\hfill\(\triangleright\) \textit{#1}}

\title{WorldDirector: Building Controllable World Simulators with Persistent Dynamic Memory}

\author{%
  \textbf{Hanlin Wang}$^{1,2}$ \quad
  \textbf{Hao Ouyang}$^{2}$ \quad
  \textbf{Qiuyu Wang}$^{2}$ \quad
  \textbf{Wen Wang}$^{3}$ \quad
  \textbf{Qingyan Bai}$^{1,2}$ \\
  \textbf{Ka Leong Cheng}$^{2}$ \quad
  \textbf{Yue Yu}$^{1,2}$ \quad
  \textbf{Yixuan Li}$^{4,2}$ \quad
  \textbf{Yihao Meng}$^{1,2}$ \quad
  \textbf{Zichen Liu}$^{1,2}$ \\
  \textbf{Yanhong Zeng}$^{2}$ \quad
  \textbf{Yujun Shen}$^{2}$ \quad
  \textbf{Qifeng Chen}$^{1}$\thanks{Corresponding author.} \\
  $^{1}$HKUST \quad
  $^{2}$Ant Group \quad
  $^{3}$ZJU \quad
  $^{4}$CUHK
}

\begin{document}

\maketitle

\begin{center}
    \vspace{-25pt}
    \centering
    \includegraphics[width=0.95\linewidth]{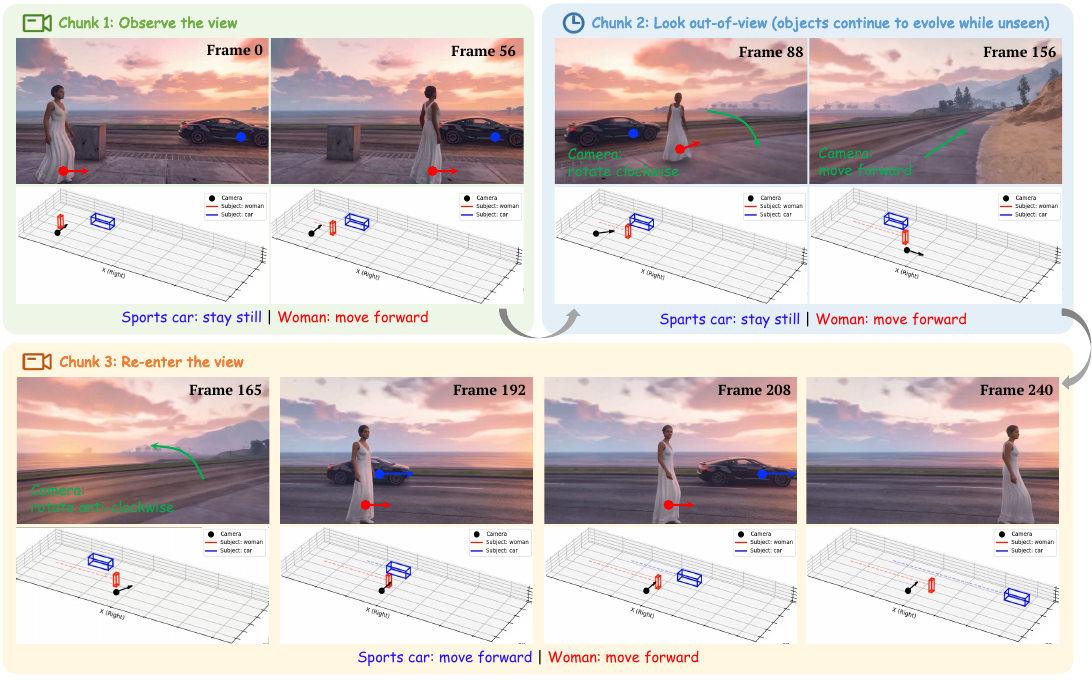}
    \captionof{figure}{\textbf{Controllable world simulation with persistent dynamic memory via \method.} By decoupling 3D semantic orchestration from latent video synthesis, our framework autoregressively generates long-horizon videos via causal chunks, ensuring rigorous dynamic memory and object permanence. Please refer to the video results on our \href{https://worlddirector.github.io/}{project page} for intuitive demonstrations.}
    \label{fig:teaser}
\end{center}

\begin{abstract}

We present \method, a highly controllable video world model framework designed for persistent dynamic object memory and unrestricted viewpoint exploration. Unlike existing world models that entangle physical dynamics with pixel rendering and rely on continuous visual observation to sustain motion, our framework explicitly decouples semantic motion orchestration from visual generation. By leveraging an LLM to coordinate 3D trajectories with camera movements and subsequently employing these orchestrated trajectories as control signals for video generation, our approach ensures strict physical logic and appearance stability, successfully preserving the exact visual identities of dynamic entities even when they re-enter the scene after prolonged periods out of view. Experimental results demonstrate that our method supports the synthesis of complex and extended events with unprecedented controllability and persistent dynamic object memory.

\end{abstract}
\section{Introduction}\label{sec:intro}

The landscape of video generation is undergoing a profound transformation, transitioning from passive pixel synthesis ~\cite{svd, lumiere, videopoet, seine} to interactive environment simulation~\cite{sora, genie, gamegenx, matrixgame2, yume, lingbot, worldplay}. A cornerstone of this paradigm shift is memory: the ability to maintain consistency of static scenes and the continuous movements of dynamic objects, whether they are visible in the frame or out-of-view. While recent methods have achieved remarkable success in preserving static scene consistency through memory retrieval or contextual conditioning~\cite{contextasmemory, worldmem, streamingt2v, freenoise, infinite}, a crucial area remains largely unexplored: \textbf{``Object Permanence''} and \textbf{``Dynamic Object Memory''}. Specifically, this entails that dynamic entities persistently exist and execute their physical movements independent of camera visibility. Consequently, whenever the dynamic objects reappear in the frame, their newly updated positions and states should be accurately observed.

To achieve this, we argue that a world simulator with robust dynamic memory must be built upon two foundational pillars. First, entities must exhibit independent motion. Their trajectories should follow continuous physical logic unconstrained by camera visibility, ensuring that unobserved dynamics progress naturally. Second, the system must guarantee strict appearance consistency. When a hidden entity re-enters the frame, its visual identity and fine details must remain entirely intact without distortion. Satisfying these two criteria is the prerequisite for elevating unpredictable video generation to the level of persistent world simulation.
Driven by this goal, several methods have been proposed to realize world simulators equipped with dynamic memory recently. One framework~\cite{liveworld} introduces a monitor-based mechanism to address out-of-sight dynamics by registering explicit ``monitors'' that autonomously track and fast-forward the temporal progression of unobserved active entities. However, this explicit tracking system scales poorly and incurs prohibitive computational overhead in scenarios involving multiple dynamic entities. Conversely, another approach~\cite{hydra} tracks dynamic features but delegates trajectory extrapolation entirely to internal generative priors. While this implicit estimation might suffice for brief occlusions, it fails during prolonged camera diversions or intricate dynamic interactions. Relying on generative weights to guess continuous physical evolution without a dedicated orchestration mechanism inevitably leads to trajectory collapse, frozen states, or severe identity errors upon re-entry.

To overcome the aforementioned limitations and fulfill the two foundational pillars, we introduce \method. Our primary insight is to explicitly decouple the motion planning of dynamic objects from the video synthesis process. By leveraging controllable generation paradigms, we transmit semantic-level planning results as conditions to the generative model, thereby realizing a persistent world simulator equipped with robust dynamic memory. This architecture not only guarantees the independent and continuous movements of dynamic objects, but also provides high controllability, enabling users to independently dictate the specific actions and semantic behaviors of multiple distinct dynamic entities.
Specifically, we employ an LLM to act as a central orchestrator, which translates user instructions into 3D bounding box and camera trajectories. These spatial plans are subsequently projected into 2D bounding box sequences, providing location conditions for video synthesis. To prevent identity distortion when a hidden entity re-enters the frame, we propose an Appearance Binding mechanism that injects RGB dynamic object features from context as visual anchors. For granular state control, a spatial-aware cross-attention mechanism~\cite{worldcanvas} routes entity-specific text prompts to their corresponding regions. Integrated within a causal autoregressive architecture, these mechanisms ensure extended video generation with strict dynamic memory.

Extensive evaluations demonstrate that \method synthesizes highly controllable dynamic scenarios while rigorously maintaining dynamic memory across extended sequences. By ensuring object permanence and appearance consistency after prolonged out-of-view intervals, our approach transcends passive video generation and represents a significant step toward interactive and persistent world simulators with unprecedented dynamic object memory.
\section{Related Works}

\subsection{Foundation Video Models and World Simulators}

Generative video synthesis has progressed rapidly with diffusion and transformer architectures~\cite{svd, lumiere, videopoet, seine}. Beyond pixel fidelity, the field is increasingly shifting toward video world models for simulating interactive environments. Pioneering works such as Sora~\cite{sora}, Genie~\cite{genie}, Oasis~\cite{oasis}, and DIAMOND~\cite{diamond} treat generative models as rudimentary physics engines, with further advances in game-like simulators~\cite{gamegenx, matrixgame2} and long-sequence interactive generators~\cite{yume, relic, worldplay}. However, relying on generative models to implicitly memorize object states, actions, and appearances overloads their capacity: when active entities exit the camera's field of view, these entangled models fail to sustain dynamics, causing objects to freeze or vanish. Our work addresses this by explicitly decoupling semantic motion orchestration from visual rendering.

\subsection{Controllable Video Generation}

To move beyond random generation, controllable synthesis has been widely explored. Image control mechanisms~\cite{controlnet, t2iadapter, vace, ditto} have been extended to video~\cite{controlvideo}. For spatial and motion control, Boximator~\cite{boximator} and GLIGEN~\cite{gligen} leverage bounding boxes, while others target camera trajectories~\cite{cameractrl} or motion tracking~\cite{dragnuwa, motionctrl}. More recently, dense or point-trajectory guidance has emerged as a flexible interface for fine-grained, entity-level motion control~\cite{tora, draganything, motionprompting, magicmotion, levitor, wanmove}. Though effective in short clips, these methods lack the autoregressive memory required for long-horizon simulation. Our framework adopts the spatial-aware cross-attention of GLIGEN~\cite{gligen} within a persistent memory architecture, enabling consistent control across extended temporal windows.

\subsection{Memory Mechanisms in Video World Models}

Memory underpins temporal coherence beyond the immediate context window. Long video generators~\cite{streamingt2v, freenoise} use sliding windows but struggle with extended occlusion. Prior work preserves static-scene consistency via FOV retrieval~\cite{contextasmemory, worldmem} or 3D representations~\cite{vmem}, yet assumes a static world. Object permanence, defined as objects persisting and evolving when unobserved, remains a core challenge in physical reasoning~\cite{clevrer, phyre, vjepa}, and is even harder for active entities in complex scenes. Recent approaches employ implicit hybrid memory tokens~\cite{hydra} or external monitors for out-of-sight dynamic simulation~\cite{liveworld}. However, internal priors risk trajectory collapse during prolonged diversions, while external monitors are computationally prohibitive. Coupled with an Appearance Binding mechanism, our LLM-orchestrated approach delivers controllable generation and robust dynamic object memory, offering a scalable path to object permanence in world exploration.

\section{Method}
\label{sec:method}

\begin{figure}[tbp]
    \centering
    \includegraphics[width=\linewidth]{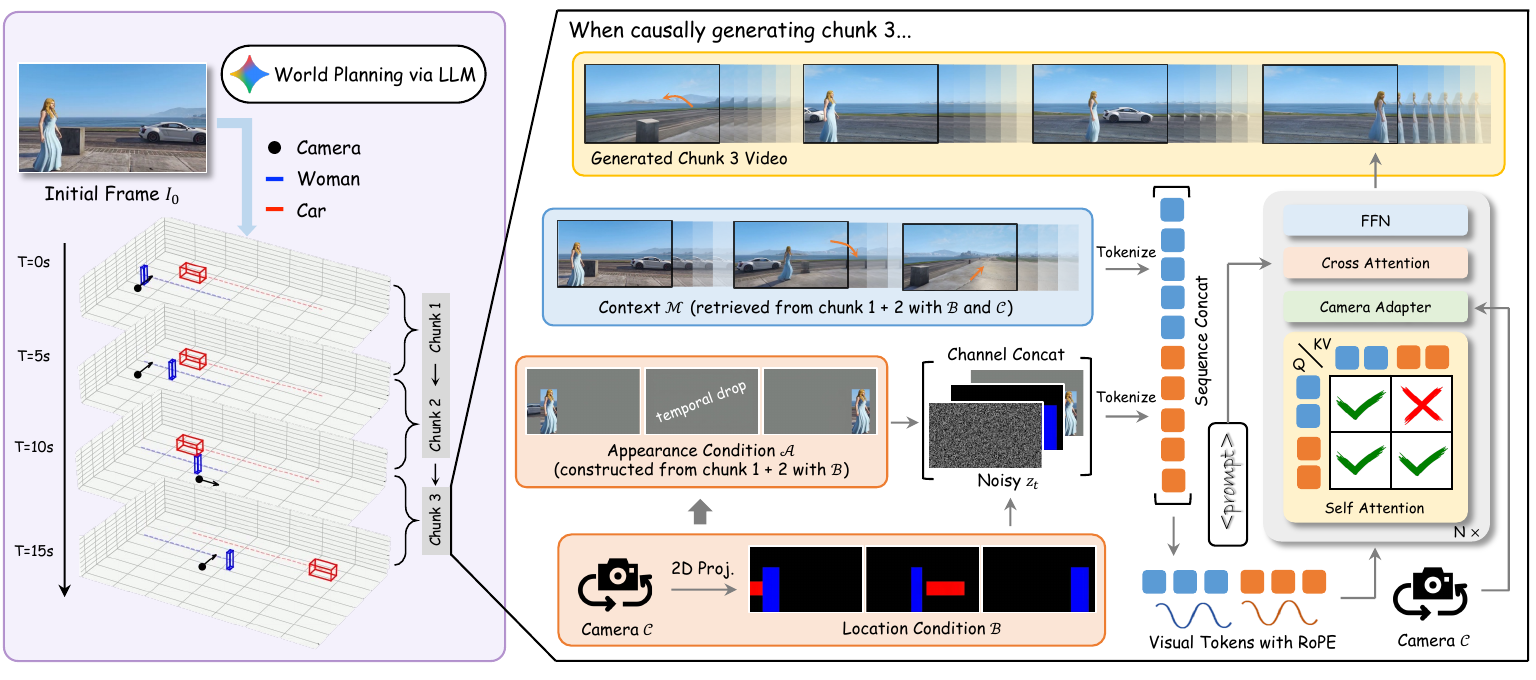}
    \vspace{-15pt}
    \caption{\textbf{Overview of \method.} An LLM orchestrates 3D trajectories that are projected into 2D Location Conditions for causal chunk generation. Location ($\mathcal{B}$) and Appearance ($\mathcal{A}$) conditions are channel-concatenated with the noisy latent, while historical Context ($\mathcal{M}$) is sequence-concatenated. During generation, \textit{temporal drop} is applied and an asymmetric attention routing prevents noise from polluting the context memory.}
    \label{overview}
    \vspace{-10pt}
\end{figure}

This section outlines our data curation pipeline (Section~\ref{data}), model design (Section~\ref{model}), training objective and inference workflow (Section~\ref{train_infer}). An overview of \method is illustrated in Figure~\ref{overview}. We use the LingBot-World-Base model~\cite{lingbot} as our foundation model.

\subsection{Data Curation Pipeline}
\label{data}
We introduce a tailored data curation pipeline to address the foundational requirements outlined in Section~\ref{sec:intro}. Specifically, this pipeline constructs comprehensive training tuples that encapsulate: 2D bounding boxes for dynamic spatial grounding, appearance references for fine-grained visual conditioning, object-centric captions detailing behavioral dynamics, and contextual signals to facilitate causal generation. The proposed pipeline comprises the following key components:

\noindent \textbf{Dynamic Object Tracking and Entity-Based Captioning.} To address the scarcity of real-world data featuring dynamic entities exiting and re-entering the field of view (FOV), we developed a game-based platform to generate 15-second videos with precise camera parameters, deliberately scripted to induce target disappearances and reappearances. We employ SAM3~\cite{sam3} to extract 2D bounding box trajectories; its robust re-identification seamlessly tracks objects despite temporary FOV exits, ensuring highly reliable annotations. For training, we sample a contiguous 5-second window from each video that maximizes the number of newly visible objects (absent in the first frame but appearing later) to specifically capture critical re-entry scenarios. The remaining 10 seconds function as a candidate pool to provide historical appearance and spatiotemporal context. Finally, we superimpose unique color-coded bounding boxes onto the source frames to preserve instance identities and feed these visually augmented sequences into Qwen2.5-VL-72B~\cite{qwenvl} to generate fine-grained textual captions of each entity's action dynamics.

\noindent \textbf{Dual-Conditioning Preparation for Dynamic Objects.} We construct two conditioning videos for each training sequence. First, to encode spatio-temporal trajectories and provide positional priors, we generate a spatial location condition video by filling each dynamic object's 2D bounding box with a unique color identifier against a zero-initialized background. Second, to ensure appearance consistency for re-entering objects regardless of their absence duration, we introduce an appearance conditioning video. Specifically, for an object $a$ at frame $t$ with bounding box $\text{box}_{a, t}$, we retrieve a reference $\text{box}_{a, t'}$ of the identical object from the 10-second candidate pool that minimizes aspect ratio divergence relative to $\text{box}_{a, t}$. The image region within $\text{box}_{a, t'}$ is then cropped, spatially resampled, and mapped onto the coordinates of $\text{box}_{a, t}$ in the appearance video, directly equipping the model with exact visual features at designated spatio-temporal indices.

\noindent \textbf{Static and Dynamic Context Retrieval.} To better support robust causal inference and preserve spatio-temporal consistency, we retrieve context through a dual-perspective approach. For static scenes, we follow ~\cite{contextasmemory} to retrieve the top-$K$ frames maximizing Field of View (FoV) overlap. For dynamic objects, we introduce a greedy algorithm prioritizing frames with active object identities within the current temporal chunk. To ensure uniform spatio-temporal distribution, we enforce a minimum temporal stride of four frames between selected frames. Ranked lists derived from both strategies are then interleaved and deduplicated to yield the final $N$ memory frame indices. The detailed selection procedure is outlined in Appendix~\ref{supp:context}.

\subsection{Building Controllable World Simulator with Persistent Dynamic Memory}
\label{model}
Building upon our established data curation pipeline, we present \method, a framework that reconceptualizes video generation as a controllable world simulation with persistent dynamic memory. We formulate this objective as a conditional denoising process guided by multi-modal structural priors. Formally, let $V \in \mathbb{R}^{T \times 3 \times H \times W}$ denote a training video sequence comprising $T$ frames. The generative process is conditioned on a composite tuple $\mathcal{T} = \{ \mathcal{B}, \mathcal{A}, \mathcal{P}, \mathcal{M} \}$:

\begin{itemize}
    \item \textbf{Location Condition} $\mathcal{B} \in \mathbb{R}^{T \times 3 \times H \times W}$ encodes the precise spatiotemporal trajectories of 2D bounding boxes for all entities, rendered as identity-preserving, color-coded masks.
    \item \textbf{Appearance Condition} $\mathcal{A} \in \mathbb{R}^{T \times 3 \times H \times W}$ provides sparse RGB features derived from contextual frames to maintain dynamic object appearance consistency across the sequence.
    \item \textbf{Multi-Granularity Prompts} $\mathcal{P} = \{p_{\text{global}}, p_1, p_2, \dots, p_k\}$ consists of a global prompt $p_{\text{global}}$ summarizing the overall video narrative, coupled with fine-grained textual descriptions $\{p_i\}_{i=1}^k$ detailing the specific semantic behaviors of $k$ dynamic entities.
    \item \textbf{Contextual Memory Frames} $\mathcal{M}$ represents contextual frames retrieved via a dual-stream selection strategy, paired with their corresponding location and appearance conditioning to align feature dimensions with the current generation window, thereby anchoring the generated content within the broader global scene.
\end{itemize}

In this section, we elaborate on how these multi-modal conditioning priors are leveraged to accomplish a dynamic memory-augmented world simulation.

\subsubsection{Control of Spatial Location and Visual Appearance}

To achieve a controllable world model with strict dynamic consistency, we extend the LingBot-World-Base architecture with auxiliary feature channels for spatial ($\mathcal{B}$) and appearance ($\mathcal{A}$) constraints, enabling high-fidelity free-exploration simulations with high physical fidelity. Specifically, $\mathcal{B}$ employs instance-specific color-coded masks to explicitly distinguish multiple entities, serving as a deterministic geometric prior for their trajectories, shapes, and orientations. Simultaneously, $\mathcal{A}$ anchors historical visual features to ensure identity coherence, preventing visual degradation when entities re-enter the camera view. 
To prevent the model from over-relying on $\mathcal{A}$ and generating unnatural sliding artifacts where entities merely translate without exhibiting proper articulated motion, we introduce a \textit{Temporal Drop Mechanism}. For each dynamic entity, we preserve a dense sequence of the initial 16 frames immediately following its entry into the view. Subsequently, we employ a sparse sampling strategy, retaining only one reference frame per six-frame interval. This information bottleneck compels the model to synthesize natural object movements driven by trajectories and semantic captions, utilizing $\mathcal{A}$ strictly as an identity anchor.

Architecturally, both $\mathcal{B}$ and $\mathcal{A}$ are encoded by a 3D VAE into latent tokens and concatenated with the noisy latent sequence along the feature dimension:
\begin{equation}
    z_{\text{in}} = \text{Conv3D} \Big( z_t \oplus \mathcal{E}(\mathcal{B}) \oplus \mathcal{E}\big(\mathcal{D}_{\tau}(\mathcal{A})\big) \Big),
\end{equation}
where $z_t$ denotes the noisy latent, and $\mathcal{E}(\cdot)$ represents the pre-trained 3D VAE encoder. $\oplus$ denotes channel concatenation. $\mathcal{D}_{\tau}(\cdot)$ formulates our \textit{Temporal Drop Mechanism}, defined as:
\begin{equation}
    \mathcal{D}_{\tau}\big(\mathcal{A}_{t}^{(i)}\big) = 
    \begin{cases} 
        \mathcal{A}_{t}^{(i)}, & \text{if } k^{(i)} < 16 \\ 
        \mathcal{A}_{t}^{(i)}, & \text{if } k^{(i)} \ge 16 \text{ and } (k^{(i)} - 16) \pmod 6 = 0 \\ 
        \mathbf{0}, & \text{otherwise}.
    \end{cases}
\end{equation}
Here, $\mathcal{A}_{t}^{(i)}$ represents the appearance condition for entity $i$ at global frame $t$, and $k^{(i)} \ge 0$ is its instance-specific relative frame index (i.e., the number of frames elapsed since entity $i$ newly entered the view). The full appearance conditioning feature $\mathcal{D}_{\tau}(\mathcal{A})$ aggregates these processed entity-level representations, where masked entities are replaced by null embeddings $\mathbf{0}$. The fused latent is then processed by a dedicated layer $\text{Conv3D}(\cdot)$, where the channel weights corresponding to $\mathcal{E}\big(\mathcal{D}_{\tau}(\mathcal{A})\big)$ are initialized from the first-frame processor for RGB identity transfer, whereas those for $\mathcal{E}(\mathcal{B})$ are zero-initialized to learn trajectory-guided generation as a residual process.

\subsubsection{Contextual Integration}

To preserve the consistency of both static scenes and dynamic objects during causal chunk generation, we integrate the retrieved context $\mathcal{M}$ through sequence-level concatenation by prepending the context frames to the noisy latent sequence along the temporal axis. To enforce structural alignment, the location and appearance conditioning associated with each context frame are concatenated along the feature dimension, strictly mirroring the input formulation of the training segment. Furthermore, to explicitly disentangle these historical anchors from the current generative chunk, the time steps used for Rotary Position Embedding (RoPE) of the context frames are shifted by an offset substantially exceeding the maximum training sequence length, establishing a definitive frequency boundary within the RoPE representation space. To prevent the noisy training latent from polluting the high-fidelity context $\mathcal{M}$, we impose an asymmetric attention mask where context tokens exclusively self-attend to remain stable, noise-free references. This allows the model to leverage historical priors without compromising contextual integrity. Finally, to equip the model with the capability to directly generate the initial sequence chunk from scratch, we randomly discard the contextual information $\mathcal{M}$ with a probability of 30\% during the training phase.

\subsection{Camera Injection and Spatial-Aware Text Control}
To accurately model perspective variations and effectively leverage contextual camera information during generation, we first convert the camera poses of all context frames and the current video chunk into relative camera poses with respect to the first frame of the current generated chunk. Following Wan~\cite{wan}, we utilize Pl\"{u}cker coordinates to encode these relative intrinsic and extrinsic parameters. Next, we apply a spatial downsampling to this representation, followed by a series of convolutional modules to extract multi-level camera motion embeddings. These embeddings are subsequently injected into each Diffusion Transformer (DiT) block via an adaptive normalization layer.

For textual condition injection, it is crucial to guarantee that entity-specific captions are precisely grounded in their corresponding spatial regions. To achieve this, we adopt the Spatial-Aware Weighted Cross-Attention mechanism from ~\cite{worldcanvas}. Rather than computing cross-attention uniformly across the entire frame, this scheme identifies the visual tokens encompassed by each entity's 2D bounding box trajectory. We then apply a targeted spatial weight bias to the pre-softmax attention logits between these localized visual tokens and the specific text tokens describing that entity. By doing so, it effectively mitigates semantic leakage and facilitates fine-grained control over multiple dynamic objects within the synthesized scene.

\subsection{Training and Inference}
\label{train_infer}
We follow the flow matching framework~\cite{flow, flow2} to perform post-training using the mean squared error (MSE) loss. The training objective is applied exclusively to the current target segment, while the historical context remains non-noisy and serves solely as a reference.
Formally, let $x_1$ denote the ground-truth latent of the target video chunk and $x_0 \sim \mathcal{N}(0, I)$ be the random noise. At a sampled timestep $t \in [0, 1]$, the training input for the target portion is $x_{tgt, t} = tx_1 + (1-t)x_0$, with the corresponding ground-truth velocity defined as $v_t = x_1 - x_0$. As described in Section ~\ref{model}, the model $u$ receives a concatenated sequence $[x_{ctx}, x_{tgt, t}]$, where $x_{ctx}$ represents the clean context tokens. The training objective is formulated as:
\begin{equation}
\mathcal{L} = \mathbb{E}_{x_0, x_1, t, \Omega} \left[ \sum_{i \in \mathcal{I}_{tgt}} \left\| u(x_t, t, \Omega; \theta)_i - v_{t, i} \right\|^2 \right],
\label{eq:flow_matching_loss}
\end{equation}
where $\Omega = \{ \mathcal{B}, \mathcal{A}, \mathcal{P}, \mathcal{M} \}$ is the union of all location, appearance, text, and contextual conditions, and $\mathcal{I}_{tgt}$ denotes the set of token indices belonging to the current video segment. By restricting the loss calculation to $\mathcal{I}_{tgt}$, we ensure that the model learns to synthesize new content anchored by the clean memory of previous frames without attempting to reconstruct the already-determined context.
During inference, our method operates in two primary stages as described below: \textit{World Planning via LLM} and \textit{Causal Chunk-Based Generation}. Further details regarding the inference implementation are provided in Appendix~\ref{supp:inference_details}.

\textbf{World Planning via LLM.} We first estimate the 3D bounding boxes of target dynamic objects in the given initial image to provide a foundational spatial context for the LLM, which then forecasts continuous 3D box trajectories---comprising both spatial coordinates and orientations---alongside our designed camera path. This trajectory planning encompasses not only the entities present in the initial frame but also those that appear later. Objects absent from the initial frame are synthesized based on their captions when they first enter the camera view. Subsequent generations are then conditioned on these initial outputs to maintain appearance consistency. These 3D trajectories are then projected onto the 2D image plane to yield a sequence of 2D bounding boxes, formulating a spatial condition $\mathcal{B}$ that strictly aligns with the location conditioning format employed during our training phase.

\textbf{Causal Chunk-Based Generation.} To facilitate computationally efficient long-horizon world exploration, we introduce an autoregressive chunk-based generation strategy (detailed in Appendix~\ref{supp:inference_details}). The projected 2D location condition $\mathcal{B}$ is partitioned into contiguous temporal chunks. During the first chunk generation, the process relies exclusively on the initial reference frame for the appearance condition $\mathcal{A}$, with an empty memory context $\mathcal{M}$. For all subsequent chunks, we recursively construct $\mathcal{A}$ and retrieve $\mathcal{M}$ from the continuously updated pool of previously generated chunks. This causal loop explicitly preserves entity identities and spatiotemporal consistency throughout the dynamic simulation, ultimately facilitating arbitrary-length world exploration.
\section{Experiments}

\noindent{\textbf{Implementation Details}} We build \method on the pre-trained LingBot-World-Base model~\cite{lingbot}. All training videos are pre-processed to a fixed resolution of $832 \times 480$ pixels at 16 fps. For conditioning, the context length is set to $N=10$ frames, with each frame independently encoded via the pre-trained 3D VAE. Our model is trained for 3,000 steps utilizing a global batch size of 64 and a constant learning rate of $1 \times 10^{-5}$. During inference, we leverage Gemini~\cite{gemini} as the orchestrator to plan 3D trajectories and states for all dynamic entities. Subsequently, the full-length video is partitioned into five-second segments and generated chunk by chunk in an autoregressive manner. Comprehensive prompt templates for the LLM are detailed in the supplementary material.

\noindent{\textbf{Baselines.}} We compare \method with state-of-the-art causal interactive world models: Yume 1.5~\cite{yume1.5}, which uses uniform temporal downsampling for memory; HY-World 1.5~\cite{hyworld1.5}, applying FOV-based attention on mixed data to achieve memorization; Infinite World~\cite{infinite}, which achieves memorization through hierarchical context compression; LingBot-World-Fast~\cite{lingbot}, leveraging causal attention for infinite generation; and HyDRA~\cite{hydra}, which utilizes spatiotemporal retrieval for maintaining off-screen character motion.

\noindent{\textbf{Evaluation Protocol.}} To evaluate our method, we use our data pipeline to construct a test set of 100 video samples featuring novel scenes and subjects that are unseen during training.
Following HyDRA~\cite{hydra}, we evaluate our model using PSNR, SSIM, and LPIPS to measure overall reconstruction fidelity via pixel-wise analysis, along with VBench's~\cite{vbench} Subject and Background Consistency for frame-level coherence. We also adopt Dynamic Subject Consistency (DSC) by cropping YOLO-detected bounding boxes of dynamic objects and computing their average DINO and CLIP similarities with their contextual counterparts. This metric effectively captures dynamic object consistency, especially for off-screen reappearance.

\subsection{Comparisons}
\label{compare}
\noindent\textbf{Quantitative Results.}
\begin{table}[tb]
  \scriptsize
  \caption{\textbf{Quantitative results}. The best and runner-up are in \textbf{bold} and \underline{underlined}.}
  \label{tab:evaluation}
  \vspace{0.5pt}
  \centering
  \resizebox{\textwidth}{!}{%
  \renewcommand{\arraystretch}{1.2}%
  \begin{tabular}{p{1.6cm}ccccccccccc} %

\toprule
\multirow{2}{*}{Method} 
    & \multirow{2}{*}{\parbox{1.3cm}{\centering \textbf{PSNR}\textuparrow}} 
    & \multirow{2}{*}{\parbox{1.3cm}{\centering \textbf{SSIM}\textuparrow}} 
    & \multirow{2}{*}{\parbox{1.3cm}{\centering \textbf{LPIPS}\textdownarrow}} 
    & \multirow{2}{*}{\parbox{1.5cm}{\centering \textbf{Subject Consistency}\textuparrow}}
    & \multirow{2}{*}{\parbox{1.5cm}{\centering \textbf{Background Consistency}\textuparrow}}
    & \multirow{2}{*}{\parbox{1.3cm}{\centering \textbf{DSC\_DINO}\textuparrow}} 
    & \multirow{2}{*}{\parbox{1.3cm}{\centering \textbf{DSC\_CLIP}\textuparrow}} 
\\
\\ %
    \hline
    Yume1.5 & 14.391 & \underline{0.455} & 0.425 & 0.898 & \underline{0.919} & 0.765 & 0.898 \\
    HY-World & \underline{14.782} & 0.418 & \underline{0.398} & \underline{0.923} & \textbf{0.931} & 0.758 & 0.911 \\
    Infinite-World & 14.574 & 0.431 & 0.406 & \textbf{0.934} & 0.908 & \textbf{0.773} & \underline{0.913} \\
    LingBot-World & 14.116 & 0.409 & 0.412 & 0.887 & 0.911 & 0.736 & 0.891 \\
    HyDRA & 13.421 & 0.352 & 0.439 & 0.855 & 0.902 & 0.632 & 0.877 \\
    \hline
    Ours & \textbf{18.127} & \textbf{0.502} & \textbf{0.359} & 0.891 &  0.909 & \underline{0.769} & \textbf{0.917} \\
   \bottomrule
\end{tabular}%
}
\vspace{-5ex}
\end{table}

As reported in Table~\ref{tab:evaluation}, \method achieves state-of-the-art performance across all three reconstruction metrics. This stems from our location conditioning, which captures continuous object positions and reflects camera poses, facilitating more accurate generation that aligns with the ground truth.
For the VBench results, Yume, HY-World, and Infinite-World attain the best performance. However, analyzing the generated videos indicates that this is largely because they generate less subject or camera motion, giving them an inherent advantage when calculating these metrics. Even though these methods also have an inherent advantage on the DSC metric due to their limited motion, our method still attains superior results. This proves our method's strong capability in preserving dynamic consistency while producing highly dynamic generations. 

\begin{figure}[tbp]
    \centering
    \includegraphics[width=0.95\linewidth]{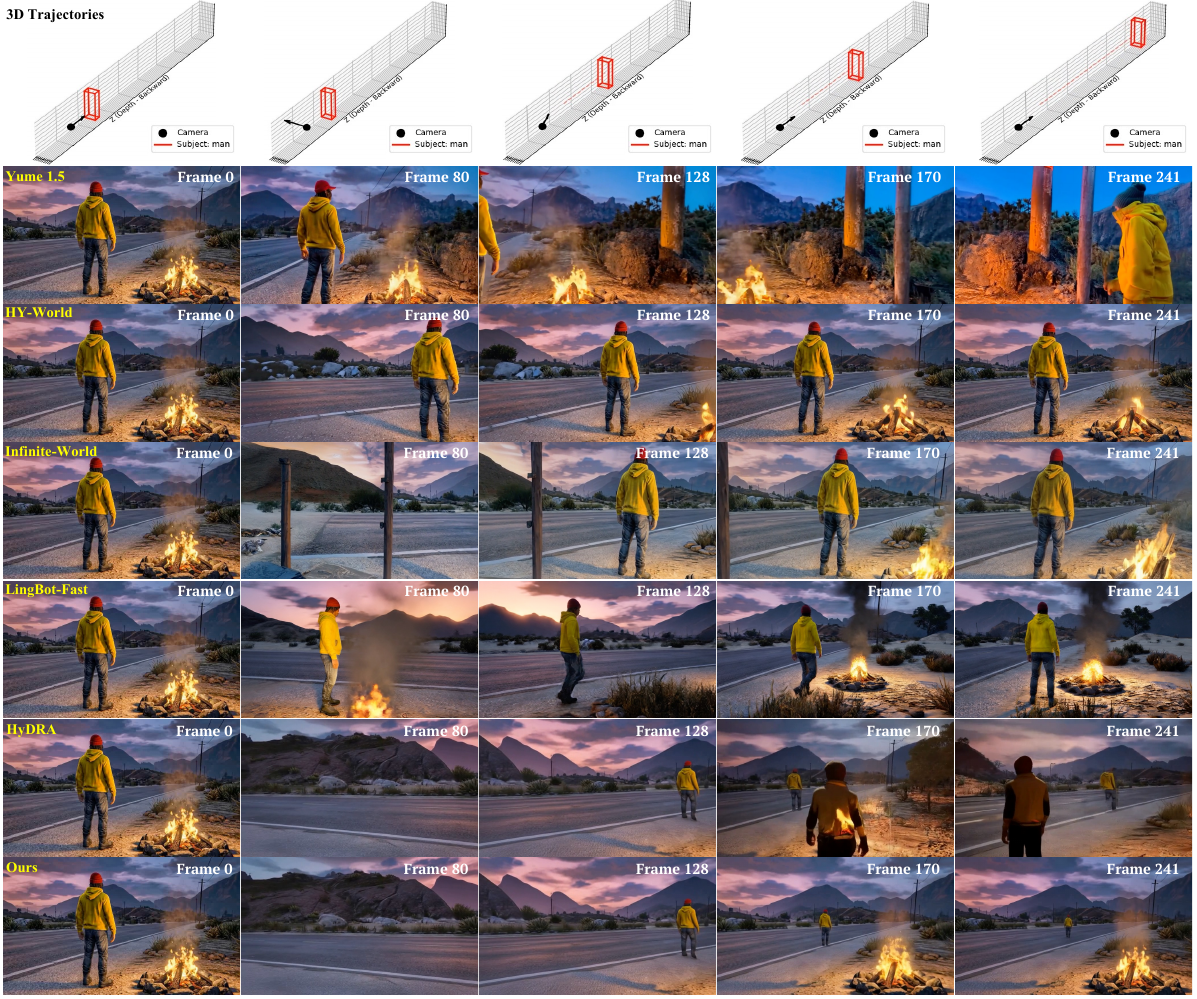}
    \caption{\textbf{Qualitative comparison with baselines.} Note that HyDRA uses the initial 10s of our results as a reference video for its generation. Please refer to the video results on our \href{https://worlddirector.github.io/}{project page} for intuitive demonstrations.}
    \label{comparison}
\end{figure}
\vspace{-1mm}
\noindent\textbf{Qualitative Results.} We show a qualitative comparison result in Figure~\ref{comparison}. Since HyDRA requires a reference video for motion extraction, we use the first 10s of our result to prompt its subsequent 5s generation. We leveraged Gemini to script a specific scenario: a man stands stationary and then walks away; concurrently, the camera pans left (moving the man out of frame) and later pans back to reveal his reappearance. Comparisons against baselines yield the following observations:
(1) Limited dynamic generation: Yume, HY-World, and Infinite-World render the man stationary even though the prompt specifies that the man walks into the distance.
(2) Identity inconsistency: While LingBot-World and HyDRA capture the man's movement, they struggle with identity preservation. Lingbot-World exhibits slight appearance degradation despite keeping the man in-frame, while HyDRA generates a completely new identity upon the man's reappearance.
(3) Insufficient control: Due to the lack of Location Condition, all baselines fail to properly synchronize camera and object dynamics with the user's design. Lingbot-World automatically generates camera translation to ensure the man remains in the shot; Infinite-World executes camera controls correctly but misses object motion; HyDRA directly ignores the man in the distance and generates a new man walking in front of the camera.
In contrast, by explicitly conditioning on location and appearance conditions, our method accurately generates the user-expected scene and maintains the consistency of the man reappearing after a long period of disappearance. We show more qualitative comparison results in Appendix~\ref{supp:comparisons}.

\subsection{Ablation Studies and Promptable World Events}
\begin{figure}[tbp]
    \centering
    \includegraphics[width=0.9\linewidth]{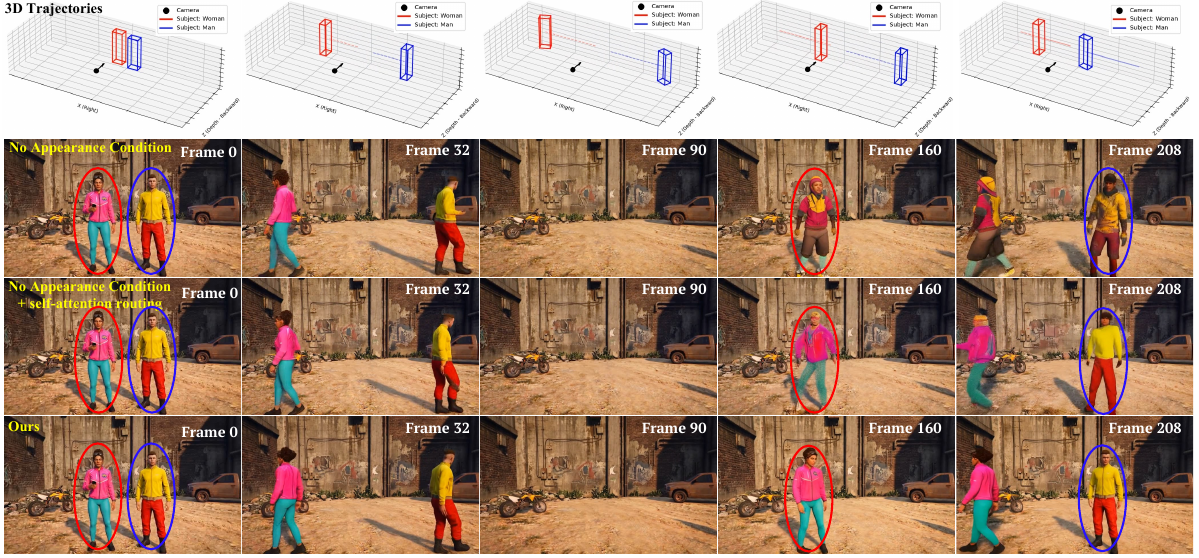}
    \caption{Ablation on Appearance Condition. We conduct experiments on a case involving complex character movements and multiple pose changes. The findings highlight the significance of the Appearance Condition for preserving dynamic consistency.}
    \label{fig:ablation1}
\end{figure}

\begin{figure}[tbp]
    \centering
    \includegraphics[width=0.9\linewidth]{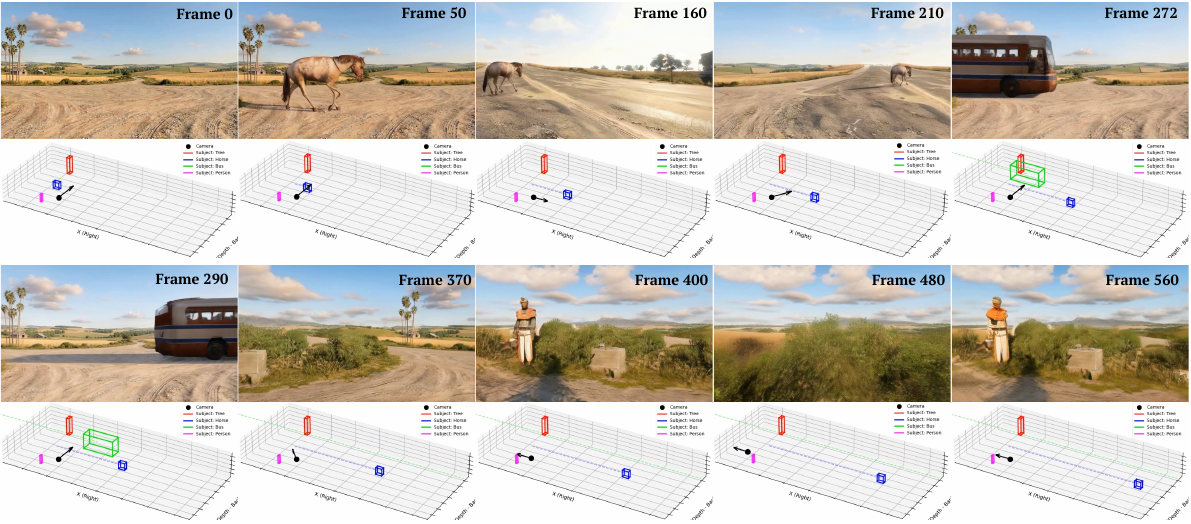}
    \caption{A generation example of Promptable World Events.}
    \label{fig:promptable_events}
\end{figure}

\begin{table}[tb]
\vspace{-6mm}
  \scriptsize
  \caption{\textbf{Quantitative ablation results on Appearance Condition}.}
  \label{tab:ablation}
  \vspace{1pt}
  \centering
  \resizebox{\textwidth}{!}{%
  \renewcommand{\arraystretch}{1.2}%
  \begin{tabular}{p{1.6cm}ccccccccccc} %

\toprule
\multirow{2}{*}{Method} 
    & \multirow{2}{*}{\parbox{1.3cm}{\centering \textbf{PSNR}\textuparrow}} 
    & \multirow{2}{*}{\parbox{1.3cm}{\centering \textbf{SSIM}\textuparrow}} 
    & \multirow{2}{*}{\parbox{1.3cm}{\centering \textbf{LPIPS}\textdownarrow}} 
    & \multirow{2}{*}{\parbox{1.5cm}{\centering \textbf{Subject Consistency}\textuparrow}}
    & \multirow{2}{*}{\parbox{1.5cm}{\centering \textbf{Background Consistency}\textuparrow}}
    & \multirow{2}{*}{\parbox{1.3cm}{\centering \textbf{DSC\_DINO}\textuparrow}} 
    & \multirow{2}{*}{\parbox{1.3cm}{\centering \textbf{DSC\_CLIP}\textuparrow}} 
\\
\\ %
    \hline
    No $\mathcal{A}$ & 16.764 & 0.469 & 0.385 & 0.878 & 0.898 & 0.693 & 0.882 \\
    No $\mathcal{A}$ + routing & 17.461 & 0.486 & 0.372 & 0.881 & 0.901 & 0.686 & 0.886 \\
    \hline
    Ours & \textbf{18.127} & \textbf{0.502} & \textbf{0.359} & \textbf{0.891} &  \textbf{0.909} & \textbf{0.769} & \textbf{0.917} \\
   \bottomrule
\end{tabular}%
}
\vspace{-5ex}
\end{table}

\noindent\textbf{Ablation Studies.} We investigate whether the model can implicitly maintain visual consistency without the explicit Appearance Condition. Assuming the unique color-coded masks in the Location Condition could guide appearance retrieval from the contextual frames, we observe that the model fails to autonomously leverage this context, causing severe identity loss for re-entering dynamic objects (Figure~\ref{fig:ablation1}, second row).
Attempting to resolve this without introducing new condition channels, we applied a heuristic self-attention routing strategy to amplify the attention weights between current and contextual dynamic object tokens sharing the same identity. Although this explicit bias captures general styles (the color of people's apparel remains consistent), it fundamentally disrupts the pre-trained latent distribution, inducing severe artifacts, blurring, and the loss of fine-grained textures (Figure~\ref{fig:ablation1}, third row).
We attribute these failures to imbalances in pixel distributions in the training data. Since static backgrounds account for the majority of pixels and dominate the MSE loss, the model struggles to implicitly learn the complex mappings required for high-fidelity consistency in small dynamic regions. Results in Table ~\ref{tab:ablation} also show that all metrics drop without the Appearance Condition. This confirms explicitly injecting Appearance Condition is necessary for dynamic memory.
We also conduct ablations on \textit{Dynamic Context} and \textit{Appearance Condition Drop} in Appendix~\ref{supp_ablation}.

\noindent\textbf{Promptable World Events.} 
Our framework is not constrained by the entities present in the initial frame. The LLM can freely populate the simulated world by defining identities, entrance timings, and 3D motion trajectories for novel objects. Upon first entering the camera view, their appearance and movements are synthesized directly from text prompts and appended to the Appearance Condition pool to ensure subsequent temporal consistency. As shown in Figure~\ref{fig:promptable_events}, this paradigm enables the simultaneous choreography of multiple emerging entities alongside unconstrained camera exploration. Consequently, rather than merely extrapolating existing video content, our approach provides a highly controllable mechanism for open-ended scene generation and dynamic environment simulation.

\vspace{-2mm}
\section{Conclusion}
\vspace{-1mm}
We present \method, a novel framework for free exploration and flexible event design in video world models while preserving rigorous dynamic memory. By decoupling semantic orchestration from latent synthesis, \method empowers LLMs to plan complex 3D trajectories and open-world events. Abstract planning is visually realized via causal chunk-based context routing, utilizing spatial and appearance conditioning. Experiments confirm our approach maintains rigorous dynamic consistency, establishing a highly controllable paradigm for future video world models.

\noindent{\textbf{Limitation.}} Relying on synthetic game data introduces a domain gap that occasionally restricts visual fidelity (e.g., unnatural locomotion or blurry faces). Future work will incorporate real-world datasets to bridge this gap and enhance overall visual realism.
\newpage
{
    \newpage
    \small
    \bibliographystyle{plainnat}
    \bibliography{main}
}
\newpage
\appendix 
\setcounter{figure}{0}
\setcounter{table}{0}

\renewcommand{\thefigure}{S\arabic{figure}}
\renewcommand{\thetable}{S\arabic{table}}

\section{Training and Compute Details.}
\label{supp:details}
In this section, we provide a more comprehensive breakdown of the training configuration and computational resources.

The training process for \method is conducted on a high-performance cluster utilizing 8 compute nodes. Each node is equipped with 8 NVIDIA A100 (80GB) GPUs, amounting to a total of 64 GPUs. To maximize memory efficiency and training throughput across this large-scale distributed setup, we employ Fully Sharded Data Parallel (FSDP) alongside activation checkpointing. This system-level optimization ensures that the memory footprint of the 3D VAE encodings, diffusion transformer blocks, and multi-modal conditioning channels is efficiently distributed, preventing out-of-memory bottlenecks when processing high-resolution video chunks and extended context memory.

As formulated in our method, the model processes training videos at a resolution of $832 \times 480$ pixels at 16 fps, with the context length explicitly set to $N=10$ frames. We optimize the flow matching objective utilizing the AdamW optimizer with a constant learning rate of $1 \times 10^{-5}$. To accelerate computation while maintaining numerical stability during the denoising process, the training is conducted using BFloat16 (BF16) mixed precision. 
Operating with a global batch size of 64, the model undergoes 3,000 optimization steps. Under the 64-GPU distributed configuration, the entire post-training pipeline takes approximately 72 hours (3 days) to fully converge.

\section{Details of Static and Dynamic Context Retrieval.}
\label{supp:context}
\begin{algorithm}[H]
\caption{Static and Dynamic Context Retrieval}
\label{alg1}
\begin{algorithmic}[1]
\Require
Candidate Context Frames $\mathcal{F}$, Training Frames $\mathcal{V}$,
         Camera Poses $\mathcal{C}$, \Statex \hspace{1.3em} 2D Bounding Boxes $\mathcal{B}$, Context Length $N$
\Ensure  Static and Dynamic Context $\mathcal{M}$

\vspace{2pt}
\Procedure{Static\_Context}{$\mathcal{F},\, \mathcal{V},\, \mathcal{C}$}
    \For{each candidate $c \in \mathcal{F}$}
        \State $\mathrm{score}(c) \leftarrow \max_{v \in \mathcal{V}}\,\mathrm{FoV\_Overlap}(\mathcal{C}_c,\, \mathcal{C}_v)$
    \EndFor
    \State \Return $\mathcal{F}$ sorted by $\mathrm{score}(\cdot)$ descending
\EndProcedure

\vspace{4pt}
\Procedure{Dynamic\_Context}{$\mathcal{F},\, \mathcal{B},\, N,\,\mathcal{V}$}
          \State Initialize coverage $\mathrm{cnt}[i] \leftarrow 0$ for each dynamic entity $i$ appears in $\mathcal{V}$
          \While{$|\mathrm{selected}| < N$}
              \State Find entity $i^* = \arg\min_i\, \mathrm{cnt}[i]$
              \Comment{least-covered entity so far}
              \State Select frame $f^* \in \mathcal{F}$ with largest $\mathrm{area}(\mathcal{B}^{(i^*)}_{f})$
              \Comment{best visible context for entity $i^*$}
              \State Add $f^*$ to $\mathrm{selected}$. Remove $f^*$ from $\mathcal{F}$
              \State $A_{\max} \leftarrow \max_{e}\, \mathrm{area}(\mathcal{B}^{(e)}_{f^*})$
              \Comment{largest bbox area in $f^*$, used as normalizer}
              \For{each entity $j$ in $f^*$}
                  \State $\mathrm{cnt}[j] \mathrel{+}= \mathrm{area}(\mathcal{B}^{(j)}_{f^*}) \,/\, A_{\max}$
                  \Comment{normalized bbox area as visibility weight}
              \EndFor
          \EndWhile
          \State \Return $\mathrm{selected}$
      \EndProcedure

\vspace{4pt}
\State $\mathcal{P}_\mathrm{cam} \leftarrow$ \Call{Static\_Context}{$\mathcal{F},\, \mathcal{V},\, \mathcal{C}$}
\State $\mathcal{P}_\mathrm{box} \leftarrow$ \Call{Dynamic\_Context}{$\mathcal{F},\, \mathcal{B},\, N$}
\State $\mathcal{M} \leftarrow [\,]$
\For{$k = 0, 1, \ldots$ \textbf{until} $|\mathcal{M}| = N$}
    \State Append $\mathcal{P}_\mathrm{cam}[k]$ to $\mathcal{M}$ if not already in $\mathcal{M}$
    \State Append $\mathcal{P}_\mathrm{box}[k]$ to $\mathcal{M}$ if not already in $\mathcal{M}$
\EndFor
\State \Return $\mathcal{M}$ sorted by temporal order
\end{algorithmic}
\end{algorithm}
In this section, we elaborate on the specific implementation details of the static and dynamic context retrieval mechanism (as outlined in Algorithm~\ref{alg1}). This algorithm aims to select $N$ memory frames from a candidate frame set $\mathcal{F}$ to construct the final context set $\mathcal{M}$. The inputs primarily include the candidate frames $\mathcal{F}$, the current training frames $\mathcal{V}$, the camera poses $\mathcal{C}$, and the 2D bounding boxes of dynamic entities $\mathcal{B}$. The retrieval process consists of two parallel scoring modules and a subsequent interleaving fusion module:

\noindent \textbf{Static Context Retrieval.} The static retrieval module (\textsc{Static\_Context}) aims to find contextual frames with the highest viewpoint overlap to provide comprehensive static background information. For each frame $c$ within the candidate set $\mathcal{F}$, the algorithm calculates the Field of View (FoV) overlap between its camera pose $\mathcal{C}_c$ and all training frame poses $\mathcal{C}_v$, taking the maximum overlap value as the candidate's score. Subsequently, all candidate frames are sorted in descending order based on these scores to generate the static context candidate list $\mathcal{P}_\mathrm{cam}$.

\noindent \textbf{Dynamic Context Retrieval.} To ensure balanced spatio-temporal coverage of dynamic objects, we maintain a coverage counter $\mathrm{cnt}[i]$ (initialized to 0) for each dynamic entity $i \in \mathcal{V}$. In each greedy iteration, we identify the least-covered entity $i^*$ and retrieve the frame $f^* \in \mathcal{F}$ that maximizes its visible 2D bounding box area. We then update the coverage for all entities $j$ present in $f^*$ by adding their bounding box areas, normalized by the maximum bounding box area $A_{\max}$ in $f^*$. This process repeats until sufficient frames are gathered, yielding the dynamic list $\mathcal{P}_\mathrm{box}$.

\noindent \textbf{Interleaving and Fusion.} We alternately append frames from $\mathcal{P}_\mathrm{cam}$ and $\mathcal{P}_\mathrm{box}$ to the final memory set $\mathcal{M}$. During this step, duplicates are discarded, and a minimum temporal stride of four frames is rigidly enforced to guarantee a uniform distribution. Interleaving terminates once $|\mathcal{M}| = N$, and the context frames are returned in chronological order.

\section{Details of Inference System}
\label{supp:inference_details}

Our inference system comprises two main components: World Planning via LLM and Causal Chunk-Based Generation. In this section, we elaborate on the specific details of these operations.

\subsection{World Planning via LLM}
We employ Gemini~\cite{gemini} as the core semantic engine for world planning. Specifically, given an initial frame, we first select the dynamic objects of interest. We leverage SAM~\cite{sam1} and DepthAnything v2~\cite{depthanythingv2} to roughly estimate the 3D bounding boxes of these objects and establish the initial orientations for both the entities and the camera. 

This structured information is then fed into the LLM, prompting it to analytically plan the corresponding 3D trajectories based on our customized narrative design. An example of the prompt we utilize is structured as follows:

\begin{quote}
\small
\textit{You are an expert with strong 3D spatial imagination capabilities. Given the following information:}

\begin{verbatim}
{
  "coordinate_system": "OpenGL (X-right, Y-up, Z-backward)",
  "camera_position": {"position": [0.0, 0.0, 0.0]},
  "camera_intrinsics": {
    "fx": 565.4046, "fy": 565.4046, "cx": 416.0, "cy": 240.0,
    "image_width": 832, "image_height": 480, "fov_v_deg": 46
  },
  "ground_height_y": [-1.248],
  "bboxes_3d": [
    {
      "bbox_3d": {
        "center": [-1.7175, -0.4237, -3.4715],
        "dimensions": [0.6799, 1.8, 0.442],
        "rotation_yaw_deg": 90,
        "prompt": "A woman walks on the road."
      }
    },
    {
      "bbox_3d": {
        "center": [3.227, -0.8014, -7.1893],
        "dimensions": [1.2749, 1.1321, 3.1872],
        "rotation_yaw_deg": 90,
        "prompt": "A car first keep still, then starts driving on the road."
      }
    }
  ]
}
\end{verbatim}

\textit{Here, "coordinate\_system" indicates the 3D coordinate system; "camera\_position" refers to the initial camera location; "camera\_intrinsics" specifies the camera parameters; "ground\_height\_y" is the y-coordinate of the ground; "bboxes\_3d" contains information for multiple subjects. For each "bbox\_3d", "center" represents the initial 3D center coordinates, "dimensions" denotes the actual width, height, and length of the object, and "rotation\_yaw\_deg" is the initial yaw angle. The "prompt" provides the textual description for the trajectory generation.}

\textit{\textbf{User Instruction:} Please help me generate the corresponding 3D bbox trajectories and camera poses for these subjects. The total duration is 15s at 16 fps. The initial camera position is at the origin, with the pose as an identity matrix facing the -Z direction.}

\textit{\textbf{0-5s:} The camera and the car remain stationary. The woman walks forward along the +X direction, reaching the edge of the camera view at 5s.}
\textit{\textbf{5-10s:} At 5-6s, the camera rotates from -Z to +X. From 6-10s, the camera moves forward along +X, overtaking the woman. The woman continues walking along +X, while the car remains stationary.}
\textit{\textbf{10-15s:} At 10s, the camera stops. From 10-11s, it rotates from +X to -Z, and from 11-15s, it remains strictly stationary. The woman walks along +X and re-enters the camera view at 11s, then continues walking within the frame. The car starts driving along +X at 10s, enters the camera view at 11s, and exits at 14s.}

\textit{Return the Python code to generate the above 3D bbox trajectories and camera poses, and visualize them. Simultaneously, for each generated frame, project the 3D bboxes onto the camera plane using the current camera pose to generate 2D bboxes. Write the projected 2D bboxes into the final output and visualize them.}
\end{quote}

Beyond planning trajectories for objects explicitly selected in the initial frame, users can also define motion paths for completely novel objects within the prompt. The LLM demonstrates remarkable capability in automatically synthesizing physically plausible kinematics for these newly introduced entities (i.e., Promptable World Events). 

Consequently, we obtain the complete 3D trajectories of all dynamic objects and their corresponding 2D bounding box sequences projected onto the camera plane. These 2D projection sequences directly serve as the deterministic Spatial Location Condition ($\mathcal{B}$) applied in the subsequent generative stage.

\subsection{Causal Chunk-Based Generation}
Building upon the projected 2D Location Condition and camera trajectories from the planning phase, we execute the video synthesis in a causal autoregressive manner, as detailed in Algorithm~\ref{alg2}.

\begin{algorithm}[t]
\caption{Causal Chunk-Based Generation}
\label{alg2}
\begin{algorithmic}[1]
\Require
Location condition sequence $\mathcal{B}_{1:T}$, Captions $\mathcal{P}$, Initial reference frame $I_0$, Total frames $T$, Chunk size $K$, Camera Poses $\mathcal{C}_{1:T}$
\Ensure  Generated continuous video stream $V$

\vspace{2pt}
\Procedure{Causal\_Generation}{$\mathcal{B}_{1:T},\, \mathcal{P},\, I_0,\, T,\, K,\,\mathcal{C}_{1:T}$}
    \State Partition $\mathcal{B}_{1:T}$ into $N = T/K$ chunks: $\{\mathcal{B}^{(1)}, \dots, \mathcal{B}^{(N)}\}$
    \State Partition $\mathcal{C}_{1:T}$ into $N = T/K$ chunks: $\{\mathcal{C}^{(1)}, \dots, \mathcal{C}^{(N)}\}$
    \State Initialize video buffer $V \leftarrow I_0$
    \For{$n = 1, 2, \ldots, N$}
        \If{$n = 1$}
            \State $\mathcal{A} \leftarrow$ \Call{Appearance\_condition\_generation}{$\mathcal{B},\,I_0$}
            \State $\mathcal{M} \leftarrow \emptyset$
            \Comment{no historical context for the first chunk}
        \Else
            \State $\mathcal{A} \leftarrow$ \Call{Appearance\_condition\_generation}{$\mathcal{B},\,V$}
            \State $\mathcal{M} \leftarrow$ \Call{Context\_Retrieval}{$\mathcal{B},\,\mathcal{C},\,V$}
        \EndIf
        \State $I_{start} \leftarrow V_{\text{last}}$
\Statex \Comment{use the last frame of buffer V as initial frame for current chunk generation}
        \State $V^{(n)} \leftarrow$ \Call{\method}{$\mathcal{B}^{(n)},\, \mathcal{P},\, \mathcal{A},\, \mathcal{M},\,\mathcal{C}^{(n)},\,I_{start}$}
        \State $V \leftarrow V \cup V^{(n)}[1:]$
        \Comment{append generated chunk without first frame to the buffer}
    \EndFor
    \State \Return $V$
\EndProcedure
\end{algorithmic}
\end{algorithm}

Given the complete spatial location sequence $\mathcal{B}_{1:T}$ and camera poses $\mathcal{C}_{1:T}$, we first partition them into $N$ sequential chunks of size $K$. The generative process maintains a global continuous video buffer $V$, which is initialized with the starting reference frame $I_0$. 

During the iterative generation, the conditioning strategy adapts based on the temporal state. For the first chunk ($n=1$), the model extracts the Appearance Condition ($\mathcal{A}$) directly from $I_0$ guided by the Location Condition $\mathcal{B}$, while the historical context memory ($\mathcal{M}$) remains empty as there is no preceding temporal information. For all subsequent chunks ($n > 1$), the framework dynamically constructs $\mathcal{A}$ and retrieves the historical Context ($\mathcal{M}$) from the previously generated video buffer $V$. This retrieval mechanism strictly leverages the location constraints $\mathcal{B}$ and camera parameters $\mathcal{C}$ to fetch precise dynamic entity identities and static background anchors.

Crucially, to guarantee temporal smoothness at the chunk boundaries, the last frame of the current buffer ($V_{\text{last}}$) serves as the conditional initial frame ($I_{start}$) for generating the next chunk $V^{(n)}$. The core diffusion model, \method, processes these multimodal conditions ($\mathcal{B}^{(n)}, \mathcal{P}, \mathcal{A}, \mathcal{M}, \mathcal{C}^{(n)}, I_{start}$) to synthesize the current segment. Finally, we append the generated frames to $V$—excluding the overlapping first frame to prevent redundancy—thereby progressively unrolling the long-horizon video stream without inherent length limitations.

\section{Ablation on Dynamic Context and Appearance Condition Drop Mechanism}
\label{supp_ablation}
\begin{figure}[tbp]
    \centering
    \includegraphics[width=\linewidth]{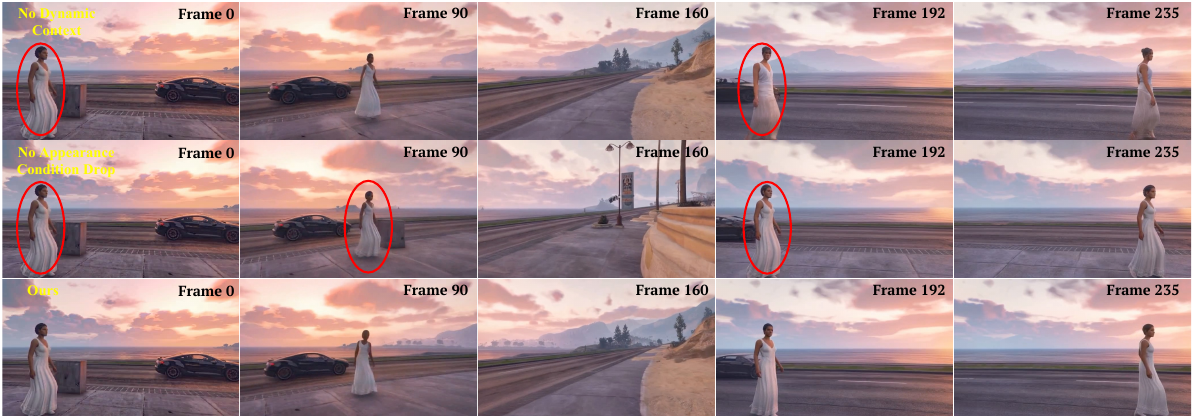}
    \caption{Ablation on Dynamic Context and Appearance Condition Drop Mechanism. We highly recommend viewing the video results on our \href{https://worlddirector.github.io/}{project page} for a more intuitive demonstration.}
    \label{fig:ablation2}
\end{figure}

We further evaluate the efficacy of the retrieved dynamic context and the Temporal Drop Mechanism. Despite the strong visual priors from the Appearance Condition, ablating the dynamic context stream confirms the necessity of retrieving dynamic objects within the contextual memory. As shown in Figure~\ref{fig:ablation2}, relying solely on Appearance Condition for re-entering dynamic entities degrades identity preservation; the model generates semantically similar but non-identical entities. This demonstrates that dynamic context is indispensable for temporally anchoring the specific object identity across causal chunks.
Furthermore, we validate the necessity of the Temporal Drop Mechanism. Removing this exposes the network to dense, frame-by-frame appearance references, which induces severe motion rigidity (e.g., characters "sliding" rather than walking naturally, as depicted in Figure~\ref{fig:ablation2}). This evidence substantiates our design: the Temporal Drop Mechanism effectively prevents overfitting to static reference images, compelling the model to synthesize fluid, text-driven dynamics rather than executing rigid image warping.

\begin{figure}[tbp]
    \centering
    \includegraphics[width=\linewidth]{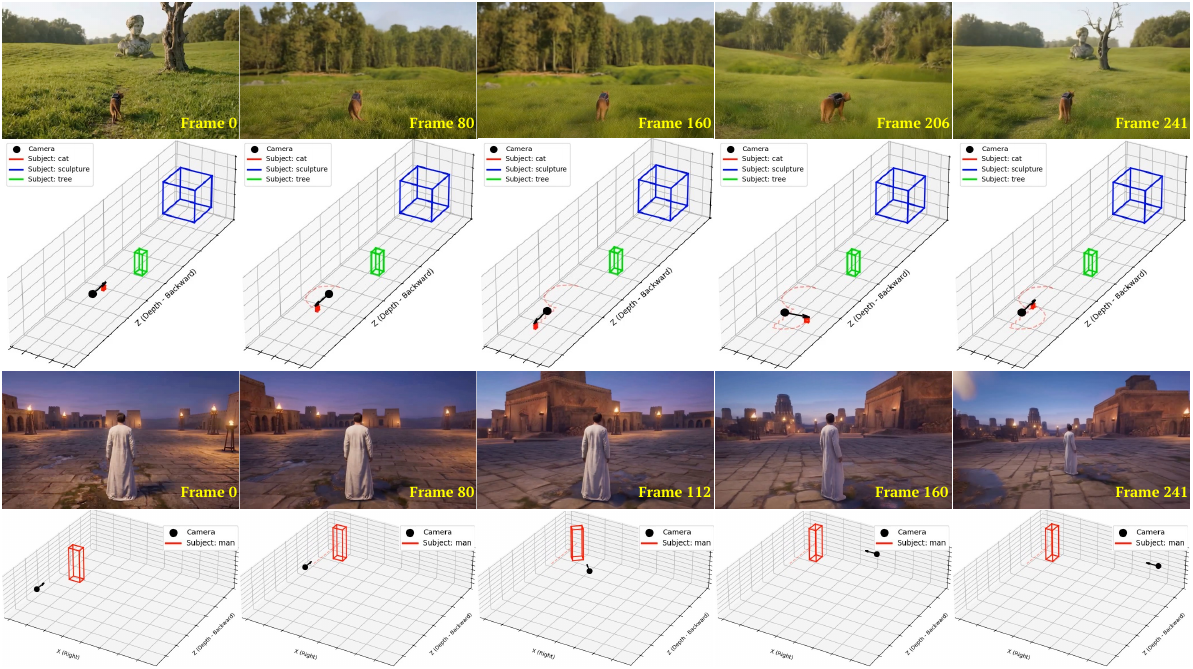}
    \caption{\textbf{Flexible Viewpoint Control.} \method supports diverse exploration paradigms. Top: A pure third-person view tracking a running dog with a $360^\circ$ panoramic sweep. Bottom: A dynamic viewpoint switch from a third-person tracking shot to an independent first-person backward movement.}
    \label{viewpoint}
    \vspace{-10pt}
\end{figure}
\section{Flexible Viewpoint Control}
\label{supp:viewpoint}
By explicitly incorporating the spatial location condition, our framework intrinsically supports flexible viewpoint control, enabling seamless transitions between first- and third-person exploration paradigms. Specifically, during 3D trajectory planning, anchoring the 2D bounding box of a target dynamic entity near the center of the camera's field of view yields a third-person perspective. Conversely, decoupling the camera trajectory from dynamic objects allows for independent first-person navigation.
As illustrated in Figure~\ref{viewpoint}, the first scenario demonstrates pure third-person exploration, where the camera follows a running dog while simultaneously performing a continuous $360^\circ$ panoramic sweep of the surrounding scene. The second scenario highlights dynamic viewpoint switching within a single sequence: the initial two temporal chunks maintain a third-person perspective following a human character, whereas the third chunk smoothly transitions to a first-person view as the camera detaches and moves backward. These capabilities underscore the model's profound flexibility in directing simulated environments.

\section{More Qualitative Comparisons}
\label{supp:comparisons}

\begin{figure}[tbp]
    \centering
    \includegraphics[width=\linewidth]{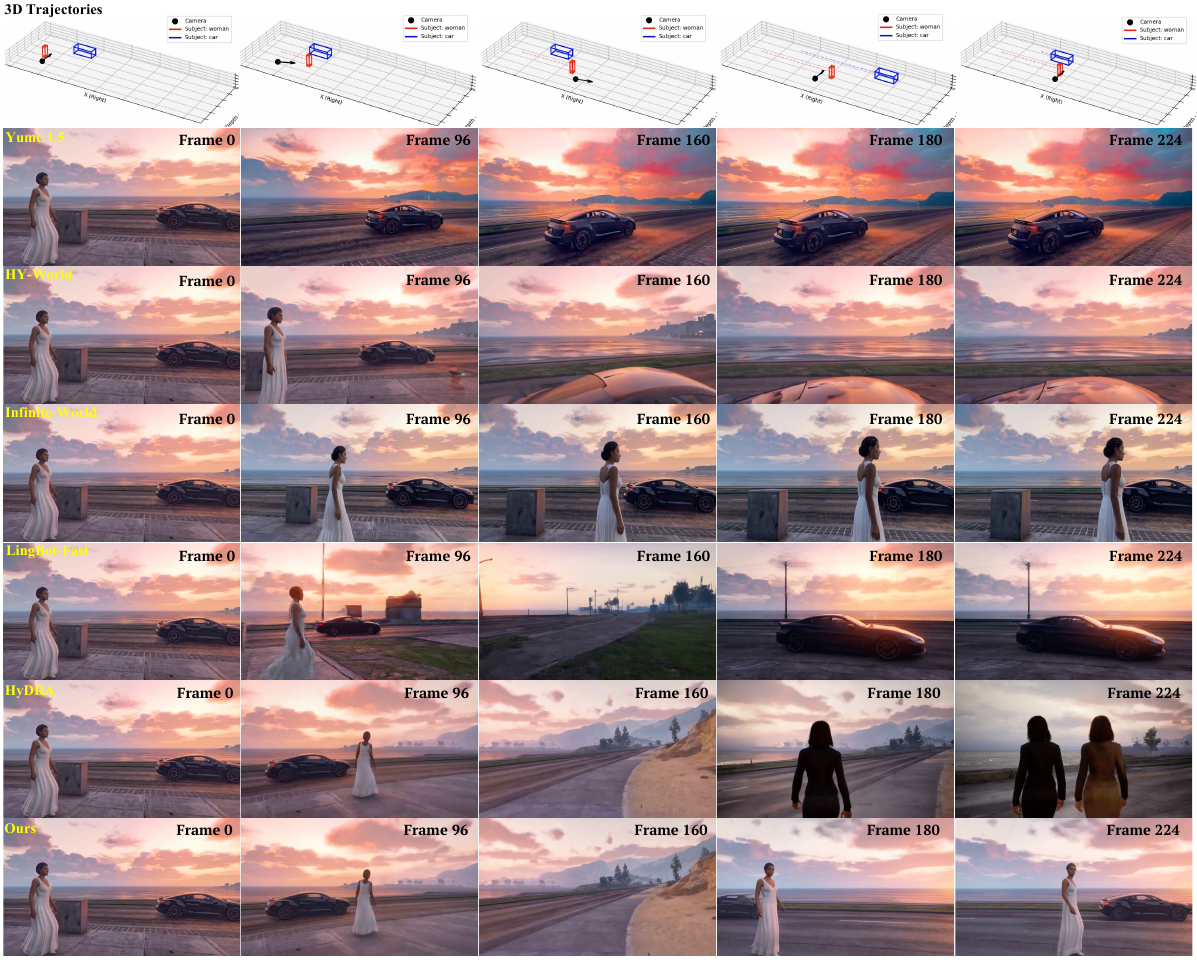}
    \caption{\textbf{Qualitative comparison with baselines.} Note that HyDRA uses the initial 10s of our results as a reference video for its generation. We highly recommend viewing the video results on our \href{https://worlddirector.github.io/}{project page} for a more intuitive demonstration.}
    \label{comp2}
    \vspace{-10pt}
\end{figure}

\begin{figure}[tbp]
    \centering
    \includegraphics[width=\linewidth]{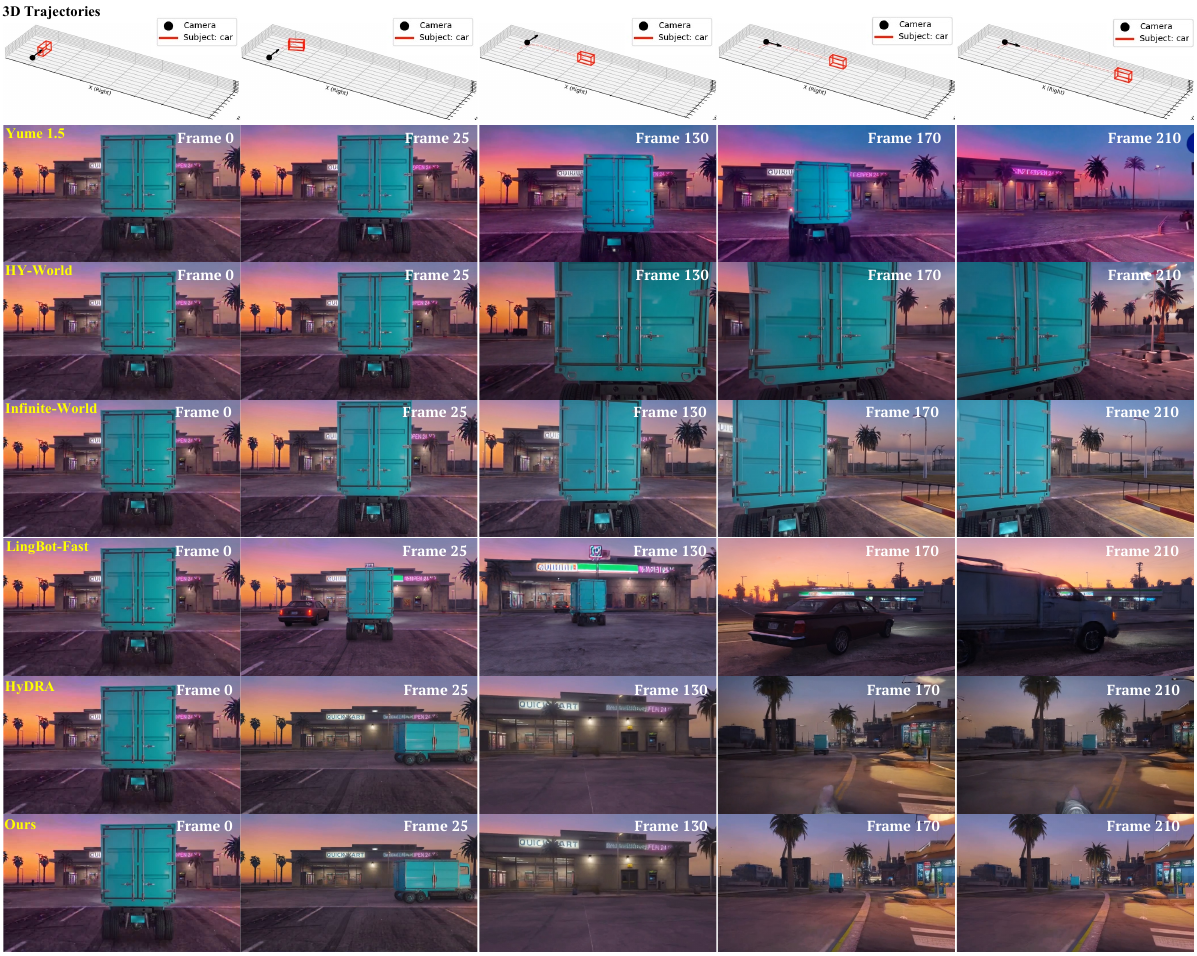}
    \caption{\textbf{Qualitative comparison with baselines.} Note that HyDRA uses the initial 10s of our results as a reference video for its generation. We highly recommend viewing the video results on our \href{https://worlddirector.github.io/}{project page} for a more intuitive demonstration.}
    \label{comp3}
    \vspace{-10pt}
\end{figure}

We provide additional qualitative comparison results in Figure~\ref{comp2} and Figure ~\ref{comp3} to further evaluate the baselines. The observations remain consistent with our main findings in Section ~\ref{compare}. Specifically, Yume, HY-World, and Infinite-World tend to generate significantly less subject motion. LingBot-World successfully produces highly dynamic results that align well with the textual prompts. However, it lacks fine-grained interactive control precision, making it difficult to strictly match the user's specific scenario designs. HyDRA consistently exhibits a strong bias towards generating a prominent subject walking directly in front of the camera, which is likely an artifact of its training data distribution. In contrast, our method accurately executes the user's intended spatial layout and maintains precise interactive control and dynamic memory.

\section{Impact Statement}
\label{supp:impact}
This paper focuses on the technical advancements in controllable video world simulation with persistent dynamic memory. The work aims to enhance applications in virtual reality, gaming, film-making, and interactive design, which could have positive societal implications in these domains. However, this study does not directly address potential societal impacts, including possible negative consequences such as malicious or unintended uses (e.g., generating deceptive or fake video content), fairness considerations, privacy concerns, or security risks that might arise from the application of this generative technology. The paper primarily presents foundational technical research and does not discuss the commercial deployment of the technology or specific mitigation strategies for these negative impacts.

\section{Responsible Release and Safeguards}
\label{sec:responsible_release}
Because \method is a highly controllable generative video model driven by an LLM orchestrator, we plan a staged and documented release. We will release the inference codebase, LLM prompt templates, and pre-trained model checkpoints strictly intended for academic research and evaluation.

For downstream applications, we strongly recommend combining \method with established safety mechanisms that fall outside the scope of this foundational paper. These include LLM prompt safety filters to prevent malicious planning, generated-video watermarking, content provenance metadata, and deployment-time monitoring. Given that our framework facilitates the continuous generation of long-horizon events with persistent entities, it inherently lowers the barrier for creating complex, logically coherent synthetic scenarios. Therefore, these external safeguards remain an important consideration to mitigate the general risks of misuse associated with video generation technologies.
\newpage
\end{document}